\documentclass[acmtog, nonacm]{acmart}
\usepackage[T1]{fontenc}
\usepackage{graphicx}
\usepackage{amsmath}
\usepackage{booktabs}
\usepackage{array}
\usepackage{subcaption}
\usepackage{graphicx}
\usepackage{float}
\usepackage{enumitem}

\AtBeginDocument{%
  \providecommand\BibTeX{{%
    \normalfont B\kern-0.5em{\scshape i\kern-0.25em b}%
    \kern-0.8em\TeX}}
}
\fancypagestyle{acceptedpaper}{%
  \fancyhf{}
  \fancyhead[L]{%
    Accepted for publication in
    \textit{ACM SIGENERGY Energy Informatics Review},
    Volume 6, Issue 3, September 2026.%
  }
  
}

\begin{document}
%% added for preprint
\settopmatter{printacmref=false}
\renewcommand\footnotetextcopyrightpermission[1]{}
\title{Learning to Run Power Networks: Effective AlphaZero-inspired Topological Control}

\author{Lukas Zetto}
\affiliation{%
  \institution{Karlsruhe Institute of Technology}
  \city{Karlsruhe}
  \country{Germany}}
  \orcid{0009-0006-4454-6058}
\email{lukas.zetto@student.kit.edu}

\author{Benjamin Sch{\"a}fer}
\affiliation{%
  \institution{Karlsruhe Institute of Technology}
  \city{Karlsruhe}
  \country{Germany}}
\orcid{0000-0003-1607-9748}
\email{benjamin.schaefer@kit.edu}

\author{Qiong Huang}
\authornote{Corresponding author}
\affiliation{%
  \institution{Karlsruhe Institute of Technology}
  \city{Karlsruhe}
  \country{Germany}}
\orcid{0000-0002-1958-6094}
\email{qiong.huang@kit.edu}

\renewcommand{\shortauthors}{Zetto et al.}

\begin{abstract}
As the integration of volatile renewable energy sources increases the strain on modern power grids, the use of Reinforcement Learning (RL) for autonomous topological reconfiguration has emerged as a promising research field to keep strained grids stable and operational. Compared to traditional redispatching measures, topological actions offer a cheaper and more cost-effective way to manage grid congestion. However, their implementation is hindered by a vast combinatorial action space and strict operational constraints. This paper investigates the effectiveness of model-based AlphaZero-inspired approaches that utilize Monte Carlo Tree Search (MCTS) for proactive grid management. We systematically evaluate how reward functions, observation density, and search guidance influence an agent’s survivability. Our results demonstrate that the optimized AlphaZero approach achieves a peak survivability of 98.43\%, significantly outperforming the proximal policy optimization (PPO) variant. We find that conducting the MCTS without guidance from a prior learned policy or value function can enhance training efficiency, and that a straightforward binary survival reward provides more effective search guidance than complex, multi-objective functions. Our findings demonstrate that while AlphaZero is a powerful framework for topological control, pure reinforcement learning is not sufficient; rather, an effective and reliable system requires a `minimalist' integration of domain-specific heuristics, binary rewards, and a restricted observation space of line loads.
% This highlights the importance of objective alignment in tree-search-based RL controllers.
\end{abstract}

\begin{CCSXML}
<ccs2012>
   <concept>
       <concept_id>10010147.10010257.10010258.10010261</concept_id>
       <concept_desc>Computing methodologies~Reinforcement learning</concept_desc>
       <concept_significance>500</concept_significance>
       </concept>
   <concept>
       <concept_id>10010583.10010662.10010668.10010672</concept_id>
       <concept_desc>Hardware~Smart grid</concept_desc>
       <concept_significance>500</concept_significance>
       </concept>
 </ccs2012>
\end{CCSXML}

\ccsdesc[500]{Computing methodologies~Reinforcement learning}
\ccsdesc[500]{Hardware~Smart grid}

\keywords{Learning to Run a Power Network (L2RPN), Deep Reinforcement learning, Monte Carlo Tree Search (MCTS), AlphaZero, Topological control}

% \received{24 April 2026}

%% new env for Sharing Data and Software Artifacts
\newenvironment{datamaterial}%
{ \vspace{-0.15cm}%
    \small\noindent{\bfseries Availability of Data and Material:}\par%
    \noindent\ignorespaces}%
{ \par\noindent%
\ignorespacesafterend }%

\maketitle
%% add header for preprint 
\pagestyle{acceptedpaper} 
\thispagestyle{acceptedpaper}

\begin{datamaterial}
The data and code used in this paper are available at \url{https://github.com/KIT-IAI-DRACOS/AlphaZero_L2RPN.git}
\end{datamaterial}

\section{Introduction}
\label{sec:introduction}
% revised
The global transition toward carbon neutrality has catalyzed a fundamental shift in power system dynamics. While essential for reaching climate goals, the integration of volatile renewable energy sources (RES), such as solar and wind, introduces two main challenges to the grid: First, these generators are volatile, threatening the balance of supply and demand and thereby the stability of the grid. Second, generation is often located far from consumption centers, as seen with offshore wind farms, increasing the strain on the transmission system and leading to congested and potentially overloaded lines. These conditions can trigger cascading failures and catastrophic blackouts, paralyzing critical infrastructure as recently demonstrated in Spain and Portugal \cite{young2025}.

Conventionally, the use of Reinforcement Learning (RL) for autonomous topological reconfiguration has emerged as a promising research field to keep strained grids stable and operational. Historically, grid congestion has been managed through redispatching, which is adjusting the setpoints of generators \cite{dorfer2022power}, or curtailing renewable energy generation or feed-in management \cite{doornhof2026renewable, smard2024}. While effective, these methods are economically costly and often underutilize existing infrastructure. Topological actions, such as busbar reconfiguration and line switching, offer a more cost-effective alternative by rerouting power through underloaded sections of the grid. However, the practical implementation of topological control is hindered by the combinatorial explosion of possible grid configurations, which exceeds the capabilities of traditional exhaustive search or manual intervention \cite{lehna2023managing}.

To address this complexity, the \textit{Learning to Run a Power Network} (L2RPN) challenge was established as a 
multi-year competition series to benchmark autonomous agents in realistic grid environments \cite{marot2020learning}. These challenges have facilitated a paradigm shift from traditional redispatching toward topological control as the primary tool for grid management. While model-free reinforcement learning (RL) agents, such as Proximal Policy Optimization (PPO), have set strong reactive baselines, they inherently lack explicit look-ahead capabilities \cite{l2rpn_baselines}. In safety-critical power systems, a purely reactive agent struggles to handle cascading thermal failures because it cannot evaluate the downstream, multi-step consequences of a topological action prior to execution. 

Consequently, there is an increasing interest shift toward model-based frameworks, specifically those inspired by AlphaZero, which couple deep neural networks with Monte Carlo Tree Search (MCTS) to proactively plan stable grid configurations \cite{dorfer2022power}. AlphaZero uniqu-ely bridges this gap by utilizing MCTS to perform systematic, non-parametric exploration of the discrete combinatorial action space, while the neural network progressively learns to guide the search, mitigating the need for computationally exhaustive full-tree rollouts. 
% expand the paragraph discussing model-free vs. model-based approaches to clarify the unique role of AlphaZero and proactively justify the benchmark scale
While real-world transmission networks encompass thousands of nodes, systematically analyzing the architectural sensitivities of these model-based controllers requires a controlled environment. In this work, we utilize the standard IEEE-14 bus system. Despite its relatively small physical footprint, its combination of substations and lines still yields a vast, non-trivial combinatorial action space of 405 raw actions (reduced to 203 via symmetry constraints). This configuration serves as an ideal, computationally viable sandbox to isolate how reward structures, search guidance, and observation configurations impact MCTS performance before deploying such frameworks to larger scales.

In addition, we provide a systematic evaluation of AlphaZero-inspired approach proposed by \cite{dorfer2022power}. While AlphaZero has revolutionized board games \cite{silver2018general}, its sensitivity to design factors in physical domains like power systems remains under-explored. We bridge this research gap by analyzing the impact of three critical design dimensions: observation density, reward shaping, and action space pruning. Furthermore, we investigate the influence of MCTS guidance variants, ranging from heuristic-based leaf evaluation to learned Q-functions, and tune hyperparameters to balance computational efficiency with grid survivability. Our findings contribute to a deeper understanding of the trade-offs between search depth and training stability in autonomous grid management and can help implement AlphaZero approaches more effectively.

% arrangement of the paper
The remainder of this paper is organized as follows: Section~\ref{sec:related} reviews the previous literature in L2RPN; Section~\ref{sec:method} details the methodology, including the implementation of the AlphaZero-inspired approach and the design choices evaluated; Section~\ref{sec:experiment} presents the experimental setup and results; Section~\ref{sec:discussion} discusses the findings and their implications; and Section~\ref{sec:conclusion} summarizes the contributions and outlines future research directions. %Our code repository is available on GitHub\footnote{\url{https://github.com/KIT-IAI-DRACOS/AlphaZero_L2RPN.git}}.

\section{Related Works}
\label{sec:related}
The application of Deep Reinforcement Learning (DRL) to power systems has been formalized by the L2RPN challenges, which have taken place annually between 2019 and 2023 \cite{marot2020learning}. These competitions utilize realistic grid environments to benchmark autonomous agents' ability to maintain stability via topological control, focusing on real-time operational windows to prevent cascading failures \cite{van2025optimizing}.

\subsection{Evolution of the L2RPN Challenge}
The inaugural 2019 IJCNN challenge utilized the IEEE 14-bus system and established the core objective of operating a grid using only bus-splitting and line-switching actions. The winning solutions from this first competition introduced several foundational concepts that have since become default standards in the field:
\begin{itemize}
    \item Action Space Reduction: The A3C-2019 solution first identified the need to eliminate redundant busbar symmetry actions, a strategy now included by default in the Grid2Op package to manage combinatorial complexity \cite{matavalam2022curriculum}.
    \item Curriculum learning \cite{bengio2009curriculum}: To improve training effectiveness, early winners employed curriculum strategies, training neural networks on simpler tasks before transitioning to full grid complexity.
    \item Activation Thresholds: The DDQN-2019 solution introduced a ``warning flag'' based on line loading levels \cite{lan2020ai}. This threshold determines when the agent should intervene versus choosing a ``do-nothing'' action, a mechanism used in all subsequent L2RPN solutions.
    \item Guided Search and Simulation: Early agents began using guided exploration and the simulation of the top N predicted actions to select the best move \cite{marot2021learning}. This established a primitive form of look-ahead that serves as the conceptual precursor to the more advanced Monte Carlo Tree Search (MCTS) employed in our work.
\end{itemize}

\subsection{Model-Free Expert System Paradigms}
The current landscape of L2RPN solutions is dominated by two primary paradigms. First, extensive research has focused on model-free RL, particularly Proximal Policy Optimization (PPO) \cite{chauhan2023powrl, manczak2023hierarchical, van2025optimizing}. These methods have been iteratively refined with heuristics and curriculum frameworks, such as the Teacher-Tutor-Junior-Senior (TTJS) framework \cite{huawei2020l2rpn}. Notable examples include the 2023 winners, La Javaness \cite{delft2023l2rpn}  and Artelys \cite{lair2023artificial}, who utilized optimized PPO variants. 

Second, several successful approaches utilize expert systems or combine them with brute-force methods. These non-RL solutions have proven to be highly competitive, achieving multiple top-three placements in recent challenges \cite{martinez2021l2rpn, alibaba2022wcci, honda2022tae}. While these findings established a robust benchmark for reactive agents, they remain confined to model-free or rule-based paradigms and do not account for the look-ahead capabilities inherent in tree-search methods.

\subsection{Transition to Model-Based Approaches}
% Table of the winner of the challenge
The limitations of reactive agents have catalyzed interest in model-based frameworks, specifically those inspired by the AlphaZero framework which utilizes Monte Carlo Tree Search (MCTS) \cite{dorfer2022power}. By simulating potential sequences of actions, these agents can proactively plan stable grid configurations. The progression of top-performing methodologies is summarized in Table~\ref{tab:l2rpn_history}. 
This approach emerged as a dominant force in recent challenge iterations, achieving third and first place respectively in the 2021 ICAPS and 2022 WCCI competitions.
\begin{table*}[ht]
\centering
\caption{Historical evolution of top-performing agents and methodologies in the Learning to Run a Power Network (L2RPN) challenges. The progression shows a shift from early model-free RL toward specialized architectures involving Curriculum Learning and model-based Tree Search.}
\label{tab:l2rpn_history}
\begin{tabular}{@{}llll@{}}
\toprule
\textbf{Year} & \textbf{Challenge Track} & \textbf{Approach / Agent Name} & \textbf{Key Methodology} \\ \midrule
2019 & IJCNN (Sandbox) & DDQN-2019 & Dueling DQN + Imitation Learning \cite{lan2020ai} \\
2020 & WCCI (Robustness) & Semi-Markov Afterstate Actor-Critic (SMAAC) & SAC-based \cite{yoon2021winning} \\
2020 & NeurIPS (Robustness) & Search with Action Set (SAS) & Evolutionary strategies (ES) \cite{zhou2021action} \\
2021 & ICAPS (Trust) & Different modes & Mix of SAS \cite{marot2022learning} \\
2022 & WCCI (Adaptability) & \textbf{AlphaZero} & MCTS \cite{dorfer2022power} \\
2023 & L2RPN Delft (Remake of 2022) & PPO Rainbow-Based & PPO-Rainbow + Heuristics \cite{sintes2024how,lehna2023managing} \\ \bottomrule
\end{tabular}
\end{table*}

While the 2022 WCCI winner successfully employed MCTS, a significant research gap remains regarding the sensitivity of such agents to specific design factors. Currently, there is a lack of publicly available codebases for AlphaZero-inspired grid controllers, and a significant research gap exists regarding the sensitivity of such model-based agents to design factors such as reward shaping, MCTS guidance methods, and observation space configurations. This paper builds upon these foundations by providing a systematic guide to effectively implementing AlphaZero controllers in safety-critical grid environments.

\section{Methodology}
\label{sec:method}
For this work, we use the Grid2op \cite{marot2020learning} framework as intended for the L2RPN competition. The experiments are conducted on the IEEE-14 Grid, a standard benchmark used in the first iteration of the challenge \cite{marot2020learning,van2025optimizing}. This grid consists of 14 substations, 20 lines, 3 transformers and 5 generators. We utilize 1004 chronics, each 8064 time steps long. To ensure generalization, performance for Rainbow-PPO and AlphaZero is measured on a held-out set of $10\%$ of the chronics.

\subsection{Non-AlphaZero Baselines}
\label{sec:non-alphazero}
We retrace the progression of L2RPN agents to establish a variation of baselines:
\begin{itemize}
    \item \textbf{DQN and PPO}: Implemented via the Stable-Baselines3 library \cite{l2rpn_baselines}. PPO is evaluated in the Redispatch (continuous) and Topology (discrete) paradigms.
    \item \textbf{Curriculum Learning (TTJS)}: Integrates a Teacher-Tutor-Junior-Senior framework \cite{sintes2024how,lehna2023managing}. The agent is initialized via Imitation Learning before transitioning to active RL. (See Appendix~\ref{app:ttjs}).
    \item \textbf{Rainbow PPO}: An optimized PPO agent utilizing the RLlib library \cite{van2025optimizing}. This includes adapted reward functions and supporting heuristics, serving as our state-of-the-art model-free baseline.
\end{itemize}

\subsection{AlphaZero and MCTS Approach}
\label{sec:alphazero}
Our primary approach adapts the AlphaZero framework, utilizing Monte Carlo Tree Search (MCTS) to navigate the grid's combinatorial topology \cite{silver2018general, dorfer2022power}. In the search tree, nodes represent critical states where grid load exceeds an activation threshold.

% add explain for choice/motivation
MCTS Guidance and Configuration: The choice of these three specific search guidance variants is explicitly motivated by the need to isolate the underlying source of control intelligence within a physical, physics-governed domain:
\begin{enumerate}
    \item \textbf{The Imitation Learning (IL) Variant (no prior guidance):} Evaluates the baseline efficacy of pure look-ahead search depth and structural discovery, independent of a learned value landscape.
    \item \textbf{The Original Heuristic Approach:} Examines the performance gains achieved by narrow, domain-specific physics heuristics that act as hardcoded rules to evaluate leaf states.
    \item \textbf{The Learned Q-Function setup:} Tests the classical AlphaZero paradigm to determine whether a parameterized neural network can effectively approximate the highly non-linear, stochastically driven boundaries of power grid stability better than hardcoded rules.
\end{enumerate}
The specific settings for thresholds and search limits are detailed in the Appendix~\ref{app:}.

\paragraph{\textbf{Action Space Ablations}}
To maintain a manageable search space, we evaluate three action reduction strategies: Symmetry (SYM), ($N-0$) \cite{subramanian2021exploring}, and ($N-1$) \cite{de2025centrally} reductions. Their impact on action space size is summarized in Table~\ref{tab:action_space}.

\begin{table}[ht]
\centering
\setlength{\tabcolsep}{12pt} 
\caption{Impact of Reduction Strategies on Action Space Size for IEEE-14}
\label{tab:action_space}
\begin{tabular}{cc}
\hline
\textbf{Reduction Method} & \textbf{Total Actions (+ Do Nothing)} \\ \hline
Complete & 405 \\
SYM & 203 \\
N-0 & 142 \\
N-1 & 82 \\ \hline
\end{tabular}
\end{table}
\paragraph{\textbf{Observation Space Configurations}}
We also evaluate four incremental observation feature sets, ranging from Minimal (line loads only) to Complete (full telemetry), as detailed in Table ~\ref{tab:observation_spaces}.

\begin{table}[ht]
\centering
\setlength{\tabcolsep}{12pt} 
\caption{Overview of Observation Space Configurations}
\label{tab:observation_spaces}
\begin{tabular}{l>{\raggedright\arraybackslash}p{4cm}}
\hline
\textbf{Configuration} & \textbf{Included Features} \\ \hline
\textbf{Minimal} & Line loads ($\rho$) \\
\textbf{Custom} & Line loads, line status, cooldowns, topology \\
\textbf{Reduced} & Line load, production, load, overflow, topology \\
\textbf{Complete} & All available \texttt{Grid2op} features including voltage and time \\ \hline
\end{tabular}
\end{table}

\paragraph{\textbf{Reward Function Configurations}}
A critical factor in MCTS efficiency is the alignment of the reward signal with the search objective. We implement six distinct reward functions categorized by their complexity and goal orientation: (1) the original \textbf{AlphaZero Reward} \cite{dorfer2022power}, (2) the binary \textbf{D3QN-2022 Survival Reward} \cite{D3QN2022}, (3) the multi-objective \textbf{D3QN-2020 Composite Reward} \cite{D3QN2020}, (4) an efficiency-based \textbf{Loss Reward} \cite{yoon2021winning}, and (5) linear safety-margin rewards (\textbf{MaxRho} and \textbf{PPO Reward}) \cite{lajavanessrepo}. Detailed mathematical formulations for each are provided in Appendix \ref{app:reward}.

\paragraph{\textbf{MCTS Guidance Variants}} To optimize the search efficiency of the AlphaZero agent, we implement three distinct methods for guiding the MCTS process: an Imitation Learning (IL) Variant, an Original Heuristic Approach, and a Learned Q-Function setup. These variants differ in how they prioritize node selection and evaluate leaf nodes. The IL Variant operates without a prior learned policy or value function, relying exclusively on search depth and transition rewards to identify surviving trajectories. This essentially treats the MCTS as a systematic discovery mechanism. Full descriptions of each approach are provided in Appendix \ref{app:guidance_variants}.

\paragraph{\textbf{Baseline Configuration}} To compare the impact of different design choices, we define a Baseline AlphaZero Configuration consisting of a $0.98$ activation threshold, a maximum of 250 MCTS simulations per step, and the original AlphaZero reward function. All subsequent ablation studies on rewards, observations, and actions are compared against this standard. Detailed parameters for search limits and safety heuristics are provided in the Appendix \ref{app:baseline_details}.

\section{Experiments}
\label{sec:experiment}
\subsection{Comparative Performance and Training Efficiency} 
\label{sec:compare}
In this section, we benchmark the AlphaZero-inspired approach against established RL paradigms to validate its competitiveness and analyze the trade-offs between survival capability and computational overhead. The performance of each agent's is assessed based on the ability to maintain grid stability over the maximum horizon of 8,064 steps. 
\paragraph{\textbf{The Model-Free Landscape}} 
As shown in Table~\ref{tab:peak}, vanilla DQN and PPO-Redispatch fail to navigate the complex grid constraints, surviving only 1.1\% and 20.7\% of the horizon, respectively. Transitioning to a topological action space (PPO Topological) doubles performance, while Curriculum Learning (TTJS) provides a significant jump to 51.2\%. These results confirm that topological reconfiguration is the superior tool for grid management.
\begin{table}[ht]
\centering
\caption{Peak performances of the different RL approaches compared to the Do Nothing agent. For the PPO Rainbow, AlphaZero Baseline and Alphzero D3QN-2022 we averaged the peaks over 4 runs. Additionally, we can also see the average standard deviation over the 101 training chronics for these.}
\label{tab:peak}
\begin{tabular}{lrr}
\toprule
\textbf{Model} & \textbf{Avg. Steps survived } & Std. dev. \\
\midrule
DQN              & 89 (1.1\%) \\
Do Nothing       & 847 (10.5\%) \\
PPO Redispatch   & 1670 (20.7\%)  \\
PPO Topological  & 3250 (40.3\%)  \\
PPO Curriculum   & 4130 (51.22\%) \\
PPO Rainbow      & 7403 (91.80\%) & 1372\\
AlphaZero Baseline        & 7486 (92.83\%) & 1323 \\
AlphaZero D3QN-2022        & \textbf{7937 (98.43\%)} & 515 \\
\bottomrule
\end{tabular}
\end{table}

\begin{figure}[ht]
  \centering
  \includegraphics[width=\linewidth]{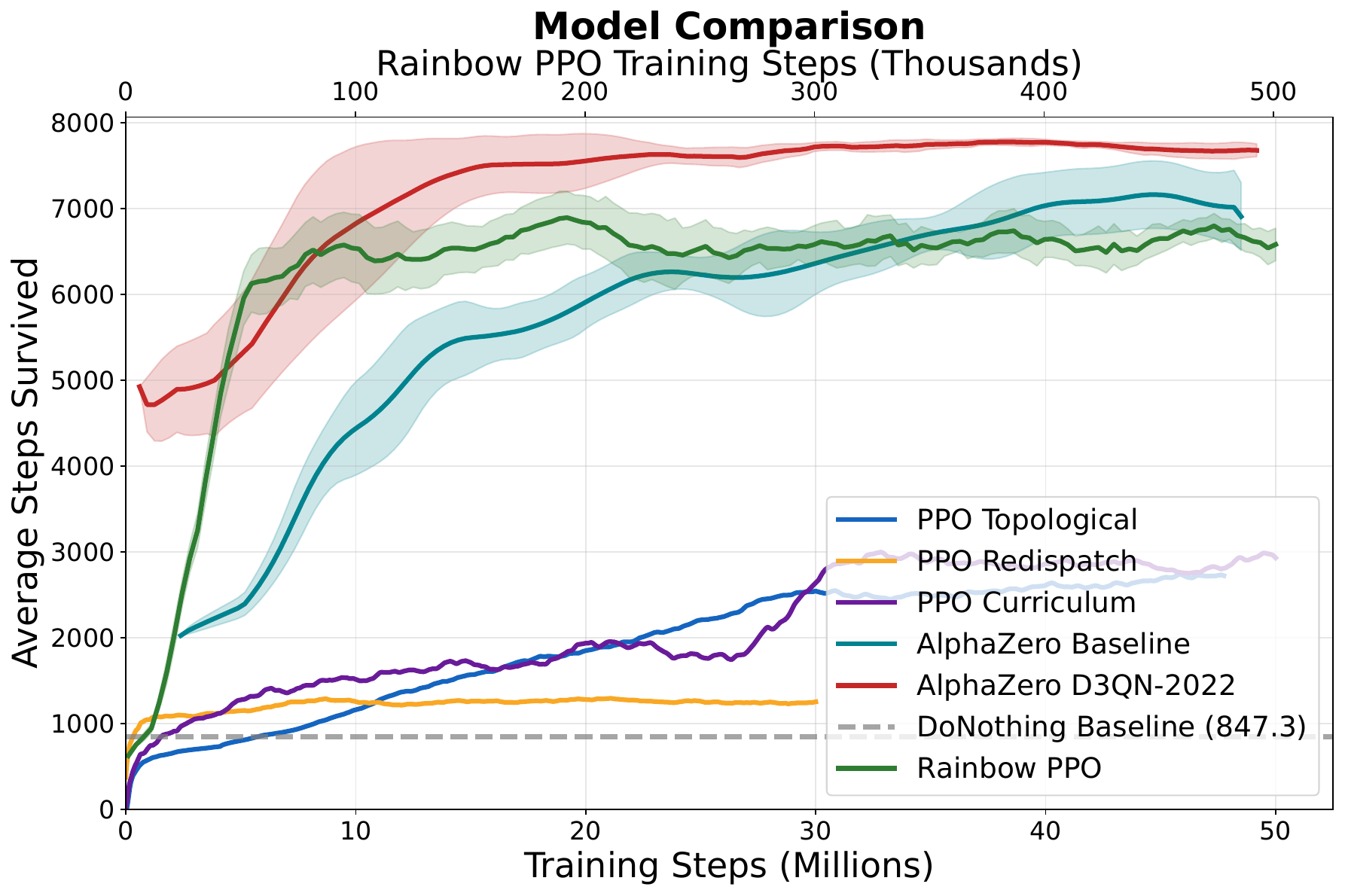}
  \caption{Comparison of AlphaZero with the different PPO variants. The results for the AlphaZero baseline described in \autoref{app:baseline_details}, the AlphaZero with the D3QN-2022 reward function as well as the PPO rainbow variant are evaluated over 4 runs and plotted with their standard deviation using a sliding window for smoothing.}
  \label{fig:methodscomp}
  \Description{}
\end{figure}

\paragraph{\textbf{The AlphaZero Advantage}}
Our AlphaZero baseline achieves 92.83\% survivability, rivaling the state-of-the-art Rainbow PPO (91.80\%). Crucially, by integrating the D3QN-2022 binary reward, AlphaZero reaches a peak survivability of 98.43\%, effectively solving the IEEE-14 benchmark.

\paragraph{\textbf{The Efficiency Trade-off}}
The average survived steps are shown in \autoref{fig:methodscomp}. The baseline AlphaZero agent achieves an average peak performance of 7486 which is comparable to the Rainbow PPO agent from \cite{van2025optimizing} which achieved 7403 steps. Although AlphaZero provides superior reliability, it requires significantly more training steps ($\sim$40M) compared to Rainbow PPO ($\sim$100k). This suggests that while MCTS is highly effective for ``running'' the network, its ``learning'' phase is computationally intensive and benefits greatly from the design optimizations explored in the next section.

As demonstrated in \autoref{tab:peak}, modifying the AlphaZero design by adopting the simple binary reward from the D3QN approach \cite{D3QN2022} significantly enhanced the performance, reaching an average peak of 7937 steps (98.43\% survivability). Beyond pure performance, this modification notably improves training efficiency, allowing the agent to converge more stably. This result not only significantly outperforms the state-of-the-art Rainbow PPO baseline but also confirms that objective alignment through sparse, survival-oriented signals is a critical factor for effective MCTS-based control. These findings provide a concrete path for improving upon the original baseline AlphaZero agent as suggested in \cite{dorfer2022power}. 

\subsection{Systematic Design Guidelines for Effective AlphaZero} 
\label{sec:desgin}
To provide actionable insights for implementing AlphaZero-inspired controllers, we perform a series of ablation studies on the core architectural components.
% This section presents the evaluation of various design configurations for the AlphaZero agent to identify optimal design aspects and architectural choices.

\begin{figure}[ht]
  \centering
  \includegraphics[width=\linewidth]{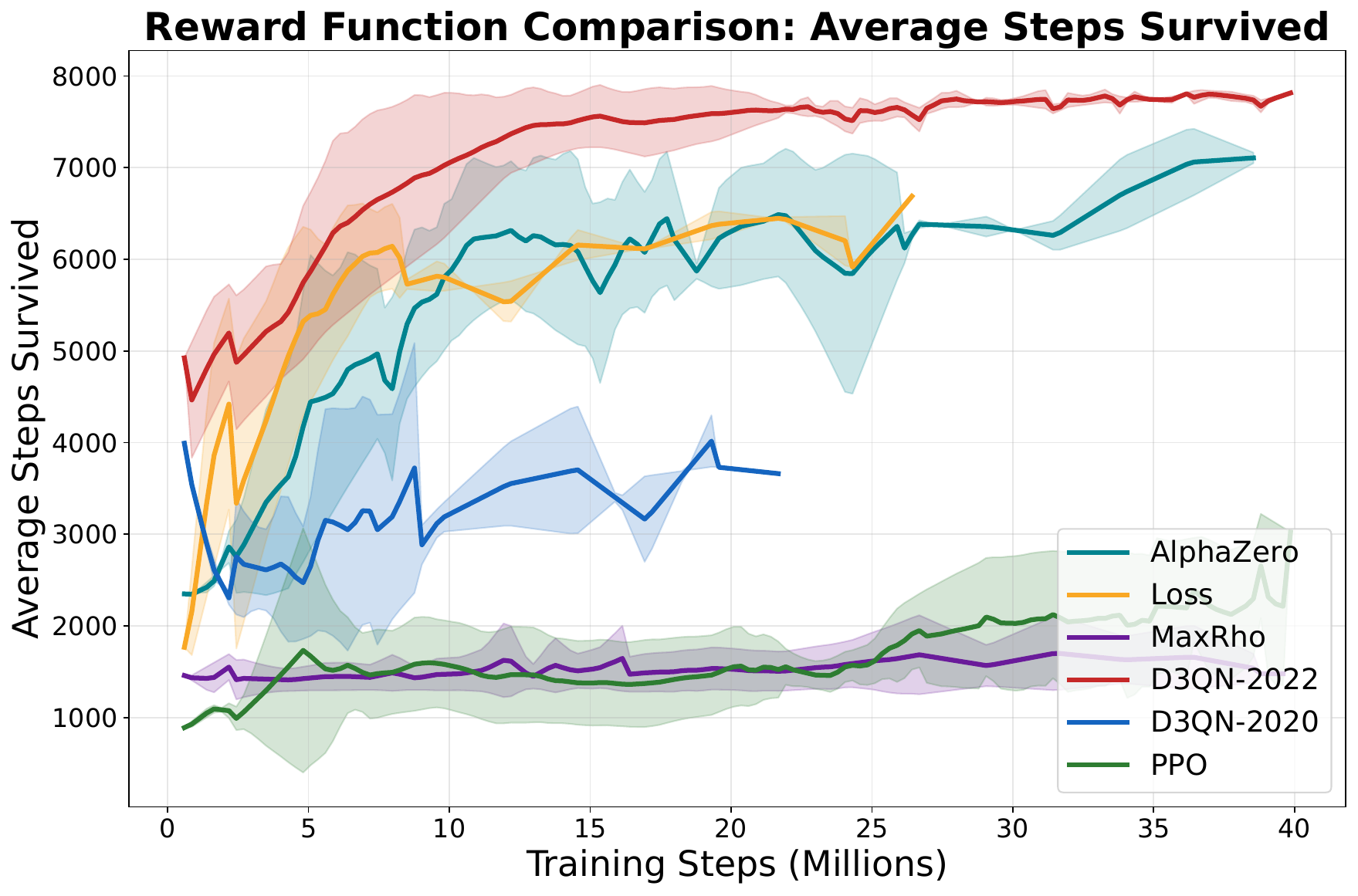}
  \caption{Training progression of the AlphaZero agent using the different reward functions presented in Appendix~\ref{app:reward}. The average performance is evaluated over 4 runs and plotted with the standard deviation using interpolation.}
  \label{fig:rewards}
  \Description{}
\end{figure}

% \subsubsection{Reward Functions}
\paragraph{\textbf{Objective Alignment (Reward Functions)}}
Contrary to the intuition that complex, multi-objective rewards provide better gradients, our results in \autoref{fig:rewards} show that he simple binary survival reward $R_{\text{D3QN-2022}}$ significantly outperforms all others, achieving a peak of 7937 steps. Complex rewards like D3QN-2020 and Loss-based efficiency signals introduced noise and slowed convergence, reaching an averaged peak performance of 4484 steps and the Loss reward showed much slower training speed as they failed to reach the 40 million training steps in the given time frame. Competitive results were reached by the originally proposed AlphaZero as well as the Loss reward as they achieved a peak performance of 7485 and 6705 respectively. 

\begin{figure}[ht]
  \centering
  \includegraphics[width=\linewidth]{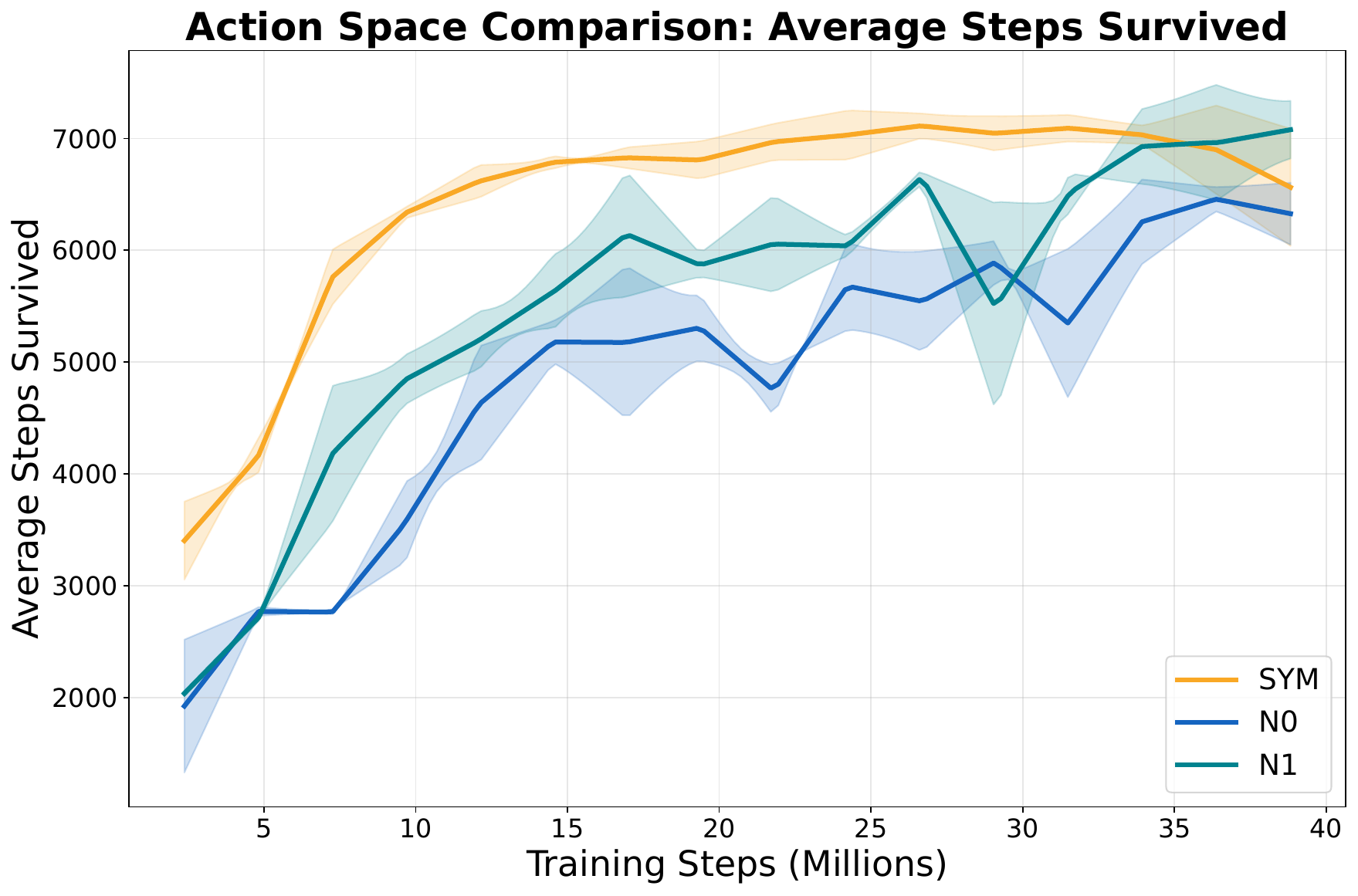}
  \caption{Training progression of the AlphaZero agent using the different action spaces presented in Section~\ref{sec:alphazero}. The average performance is evaluated over 4 runs and plotted with the standard deviation using interpolation.}
  \label{fig:action}
  \Description{}
\end{figure}

\paragraph{\textbf{Navigation of Action Space}}
\label{actions}
Surprisingly, the most restrictive reduction strategy (N-1) did not yield the best results. The Symmetry (SYM) reduction, which maintains the highest degree of topological flexibility while removing redundant actions, proved to be the fastest and most stable (\autoref{fig:action}). MCTS is robust enough to handle moderate action spaces; over-restricting the topology (e.g., N-0 or N-1) can limit the agent's ability to find ``survival paths'' in extreme congestion.

\begin{figure}[ht]
  \centering
  \includegraphics[width=\linewidth]{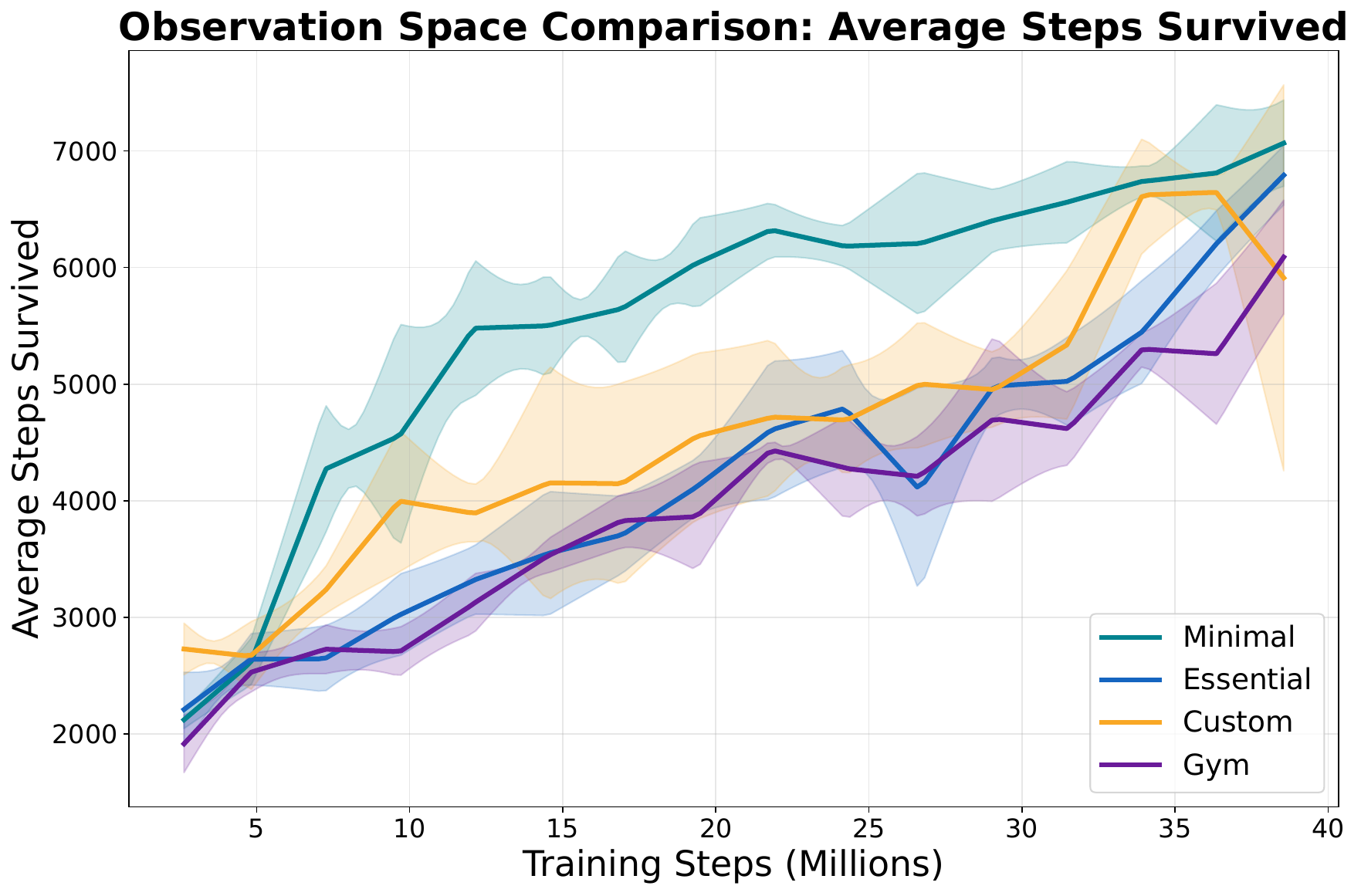}
  \caption{Training progression of the AlphaZero agent using the different observation spaces presented in Section~\ref{sec:alphazero}. The average performance is evaluated over 4 runs and plotted with the standard deviation using interpolation.} 
  \label{fig:observation}
  \Description{}
\end{figure}

\paragraph{\textbf{Information Density (Observation Space)}}
Our analysis shows that the Minimal observation space configuration, which uses only line loads, consistently outperformed configurations with complete telemetry. The result yields the fastest learning rate and the highest training stability, as can be found in \autoref{fig:observation}. Its peak performance achieved a survival time of 7340 steps. The Custom, Reduced and Complete configurations achieve similar peak performances with 7285, 7328 and 7287 steps, respectively. However, they all learn slower and are less consistent. Providing full nodal injections and voltage phase angles appeared to overcomplicate the state-action mapping, leading to slower learning and inconsistency.

\begin{figure}[ht]
  \centering
  \includegraphics[width=\linewidth]{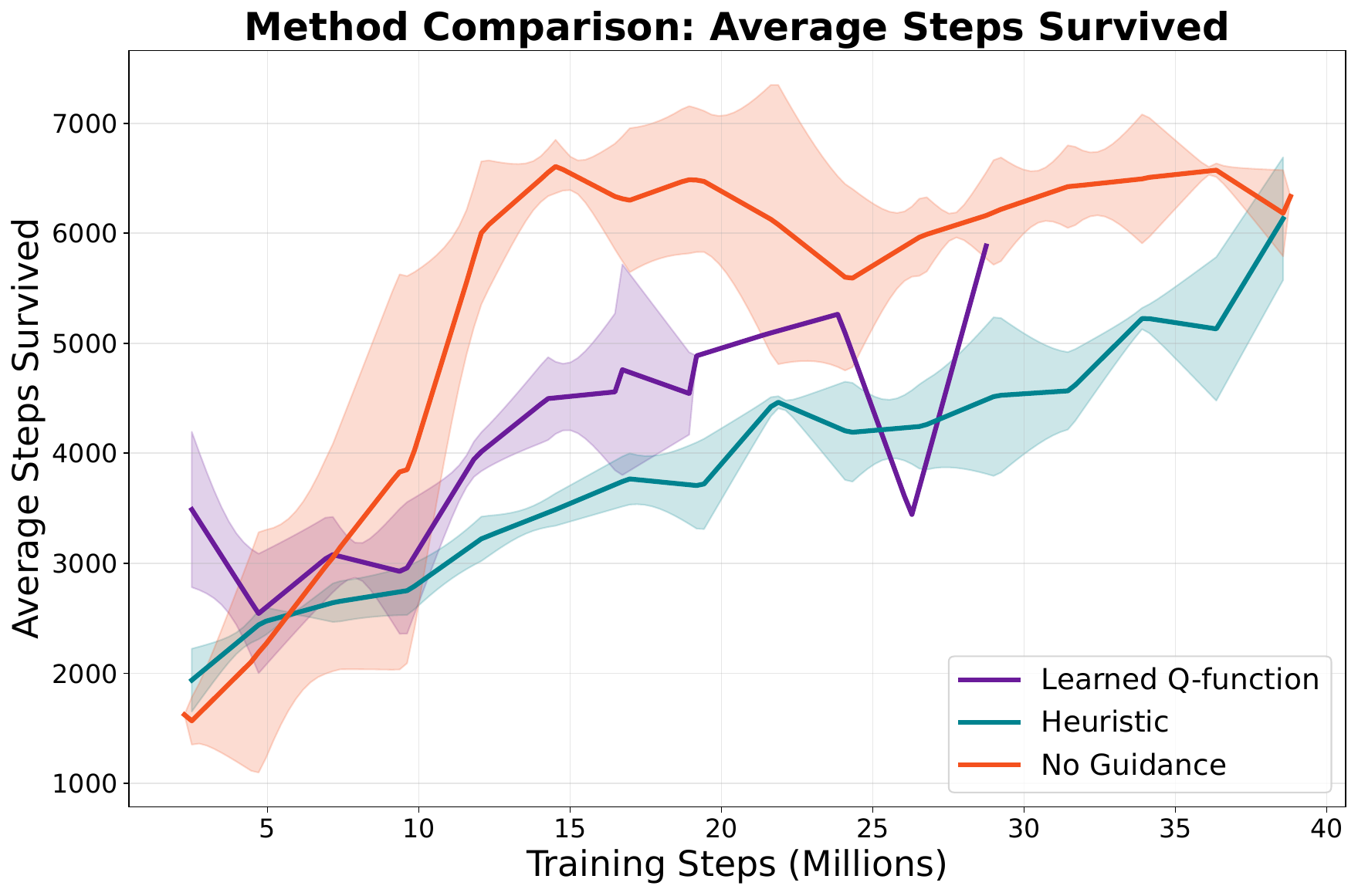}
  \caption{Comparison of the three MCTS guidance methods of using learned prior and value function, learned prior and heuristic value function or no guidance at all.}
  \label{fig:methods}
  \Description{}
\end{figure}

\paragraph{\textbf{The Guidance Paradox}}
The simplified no-guidance variant, which uses neither a prior policy nor a value-function guidance, achieved strong performance with the highest training efficiency. It reached an average peak survival of 7220 steps within only 25 million training steps. In contrast, while the average peak performance is compatible with 7250 steps, the original heuristic-based version required a lot more training steps to achieve strong performance. Finally, the integration of a learned Q-function proved to be the least effective. This configuration resulted in an extremely slow execution runtime as even with an increased runtime of 72 hours, it only reached 20-30 million training steps due to the additional overhead. While the performance for the achieved number of training steps seems competitive with the other variants, the extremely long run-time makes this approach unattractive. Unbiased MCTS search, when paired with an effective reward signal, is often more efficient than search guided by potentially inaccurate or high-latency learned functions. The results are shown in \autoref{fig:methods}.

\section{Discussion}
\label{sec:discussion}
This section analyzes the design principles for the AlphaZero inspired controller, identifying the relationship between tree-search mechanics and the physical constraints of the power system, including reward structures, action space reduction, and observation density.

\paragraph{\textbf{Optimizing Information Density (Observation Space)}}
Our findings demonstrate that AlphaZero agent performance is inversely proportional to observation complexity, with the minimal configuration (line loads only) consistently yielding the fastest convergence and highest training stability. This result challenges the intuitive assumption that denser telemetry, such as voltage phase angles, active power injections, and local demand profiles, inherently enables superior grid control. One hypothesis for this behavior is that expanding the observation space drastically increases the state-space dimensionality, introducing what can be termed as ``feature noise''. In the context of topological reconfiguration, line loading acts as the most direct physical proxy for systemic grid stress and imminent thermal failure. By saturating the neural network with granular, high-dimensional inputs, the model risks overfitting to stochastic demand fluctuations and localized noise rather than learning the structural bottleneck features critical for long-term survival.
% where the neural networks overfit to stochastic fluctuations in demand rather than the fundamental physical state of the grid.

This interpretation aligns with recent findings in Reinforcement Learning for Optimal Power Flow (OPF), where it has been shown that over-complicating or duplicating telemetry features during the training phase can yield highly counterintuitive results and degrade policy optimization \cite{butt2024reinforcement}. Rather than providing a richer context, excessive feature density complicates the state-action mapping for the neural network, ultimately impeding the value propagation within the MCTS framework. Therefore, we recommend a more conservative, simulation-verified approach to feature selection in safety-critical grid domains, prioritizing core physical constraints over complete telemetry saturation.
% In the AlphaZero experiments, we observed that the agent performed best in terms of training speed and final performance when restricted to a minimal observation space consisting only of line loads. Similar to the results in \cite{van2025optimizing} for the PPO agent, it appears that providing more granular information can introduce unnecessary noise. Since line loads are the decisive factor for grid survival, focusing the agent's attention solely on these metrics prevents the neural network from overcomplicating the state-action mapping.

\paragraph{\textbf{Objective Alignment and Sparse Signals (Reward Function)}}
Contrary to our observations regarding the observation space, the results visualized in \autoref{fig:rewards} show that the superiority of the binary survival reward ($R_{\text{D3QN-2022}}$) over complex, load-based rewards (MaxRho, D3QN-2020) highlights a critical ``Objective Alignment'' challenge in MCTS. Dense rewards attempt to guide the search via a smooth gradient, but in safety-critical environments, they often introduce conflicting subgoals, e.g., optimizing for efficiency at the cost of stability.
% purely load-based rewards (PPO Reward, MaxRho Reward) did not perform well for our AlphaZero agent. 
% While better than these, the composite reward also failed to converge to a high-quality policy, likely due to an excess of subgoals that distracted the agent from the primary objective of survival. %\newline

While the energy loss-based and the original AlphaZero reward (based on maximal line load and the number of lines that are offline) were effective, the simple binary reward proved superior in both training speed and final performance. By providing a clear ``pass/fail'' signal, the agent avoided distraction from possible subgoals. Additionally, this strategy aligns effectively with the MCTS tree algorithm, as it provides a clean, unambiguous signal for value propagation; especially since the final action selection in MCTS is purely based on the maximum steps reached within the tree. In contrast, other reward structures involving continuous or penalty-heavy metrics performed poorly, likely because they introduced a ``dense but noisy'' gradient. MCTS is a discovery-driven algorithm; if the agent is penalized for minor line-load fluctuations that do not lead to failure, the search may prune branches that contain the decisive topological maneuver needed for long-term survival. A binary signal ensures that the value propagation within the search tree remains focused solely on the terminal survival objective, preventing the distraction observed with penalty-heavy reward structures.
% In a tree-search environment, if the agent is penalized for minor fluctuations that do not lead to failure, the MCTS may prioritize safe but suboptimal paths, ultimately failing to find the decisive topological change necessary for grid survival. Consequently, binary rewards appear to be a significant improvement over the more complex, weighted rewards, including the reward originally used for the AlphaZero design.

\paragraph{\textbf{Action Space Flexibility vs. Over-Pruning}}
For the IEEE-14 grid, we found that strict action space reduction is counterproductive. The Symmetry (SYM) reduction strategy, which retains 203 actions, outperformed more restrictive $N-1$ strategies. This suggests that the combinatorial explosion of a small grid is not yet a bottleneck for MCTS. Instead, the bottleneck is the loss of reachability: over-pruning removes the unconventional busbar configurations that are often necessary to reroute flow during extreme contingencies. While larger grids will eventually require stricter dimensionality management, these results suggest that MCTS's inherent search capability is robust enough to handle moderate action spaces without losing efficiency.
% Even the symmetry-based strategy results in only 203 actions (\autoref{tab:action_space}), a size that remains manageable for AlphaZero. Over-pruning likely restricts the agent from finding optimal maneuvers, making stability harder to maintain. While these results favor larger action spaces here, stricter reductions are likely still required to manage the increased dimensionality of larger grids.

\paragraph{\textbf{The Guidance Paradox: Search vs. Prior Policy}}
The evaluation of guidance variants reveals a Guidance Paradox: the no guidance variant was the most training-efficient. In traditional AlphaZero, a prior policy is essential to prune the tree. However, in the power grid domain where state transitions are governed by physical power flow, the rewards collected directly within the search tree are often more reliable than a nascent, learned $Q$-function.
% in principle just applying a sort of IL, resulted in the fastest learning and achieved a similar peak performance to the original heuristic-guided approach. However, this IL method did seem to be slightly less stable later on. 
While a learned prior can stabilize the agent in the long run, the computational overhead of frequent neural network updates and the risk of policy collapse, where a poor early policy misguides the search, make unbiased MCTS a more effective starting point for autonomous grid controllers.

\section{Conclusion and Outlook}
\label{sec:conclusion}
This study provides a systematic evaluation of AlphaZero-inspired topological control, demonstrating that while the framework is highly capable, pure RL is not sufficient for the operational demands of modern power grids. Our results highlight that achieving a peak survivability of 98.43\% on the IEEE-14 benchmark is not a result of increased model complexity, but rather of strategic simplification and the integration of domain-specific heuristics. 

We have established a minimalist design blueprint for effective AlphaZero implementation: utilizing a minimal observation space (line loads only) and a binary survival-based reward. These choices ensure objective alignment between the neural network and the Monte Carlo Tree Search (MCTS), preventing the agent from being distracted by the ``feature noise'' and conflicting subgoals inherent in multi-objective reward structures. Crucially, this blueprint constrains the agent exclusively to topological actions. In real-world deployment, grid operators would prioritize these topological reconfigurations because they do not incur direct market costs. However, a production-grade controller would treat topology as the first line of defense to minimize expensive redispatch, falling back on generation adjustments only when topological maneuvers are exhausted.

% Additional enhancements like Curriculum Learning can improve performance and speed up training. \\
While AlphaZero successfully outperformed the state-of-the-art Rainbow PPO in survival time, this gain comes with a significant trade-off in training efficiency. The superior performance of our search-based adaptation, alongside the historical success of non-RL expert systems in the L2RPN challenges, suggests that the ``intelligence'' in grid control may lie more in the search and physics-based heuristics than in the reinforcement learning policy itself. This raises fundamental questions for the community: as we move toward larger grids, should we prioritize learning complex policies, or should we focus on optimizing the efficiency of the search and the accuracy of the underlying heuristics?

To bridge the gap between high-performance research and real-world grid deployment, future work must prioritize several key areas. First, addressing the computational overhead of MCTS is essential; integrating AlphaZero into a Curriculum Learning framework, where the agent masters small-scale topological maneuvers before tackling cascading failures, could significantly accelerate convergence. Furthermore, an Options Framework could allow agents to treat complex bus-switching sequences as temporally extended abstract actions, enabling reasoning over high-level grid states rather than individual busbar changes.

Scalability and transparency also remain primary concerns for larger networks. In general, the importance of reflecting domain knowledge in feature design has been shown for RL applications in OPF \cite{wolgast2024learning}, which provides valuable input for optimizing our minimalist observation framework in future work. Adopting Multi-Agent Reinforcement Learning (MARL) or hierarchical frameworks would allow modular AlphaZero instances to manage regional substations, reducing the global action space to a manageable level \cite{Ma1,Ma2,de2025centrally}. Finally, for autonomous agents to be adopted by grid operators, their decisions must be interpretable. Future research should prioritize Explainable RL (XRL) techniques, such as SHAP or LIME, to provide human-interpretable justifications for topological reconfigurations \cite{butt2024reinforcement,butt2024explainable}. This transparency is vital for transforming a ``black-box'' controller into a trusted decision-support tool, ultimately leading to a hybrid approach that respects the physical laws of the system through heuristics and look-ahead search while using RL to refine decision efficiency.

\begin{acks}
We gratefully acknowledge funding from the Helmholtz Association under grant No. VH-NG-1727 and the Networking Fund through Helmholtz AI. We also thank EnBW Research Department for their valuable support/discussion. The authors acknowledge support by the state of Baden-W{\"u}rttemberg through bwHPC.
\end{acks}

\bibliographystyle{ACM-Reference-Format}
\bibliography{references}

@misc{alibaba2022wcci,
  author       = {{Alibaba Research}},
  title        = {{L2RPN WCCI 2022 Competition Repository}},
  howpublished = {GitHub},
  year         = {2022},
  note         = {\url{https://github.com/AlibabaResearch/l2rpn-wcci-2022/tree/main}, Accessed 08 May 2026}
}

@misc{honda2022tae,
  author       = {{Honda Research Institute Europe}},
  title        = {{L2RPN Trial and Error (TAE) Agent}},
  howpublished = {GitHub},
  year         = {2022},
  note         = {\url{https://github.com/HRI-EU/l2rpn_tae_agent}, Accessed 08 May 2026}
}

@article{Ma1,
  author    = {E. Boguslawski and A. Leite and M. Dussartre and B. Donnot and M. Schoenauer},
  title     = {{Emulation of Zonal Controllers for the Power System Transport Problem}},
  journal   = {RJCIA},
  year      = {2024},
  volume    = {41}
}

@article{Ma2,
  author    = {G. Losapio and D. Beretta and M. Mussi and A. M. Metelli and M. Restelli},
  title     = {{State and Action Factorization in Power Grids}},
  journal   = {arXiv preprint arXiv:2409.04467},
  year      = {2024}
}

@inproceedings{de2025centrally,
  title={Centrally coordinated multi-agent reinforcement learning for power grid topology control},
  author={de Mol, Barbera and Barbieri, Davide and Viebahn, Jan and Grossi, Davide},
  booktitle={Proceedings of the 16th ACM International Conference on Future and Sustainable Energy Systems},
  pages={460--475},
  year={2025}
}

@misc{young2025,
  author       = {Holly Young},
  title        = {What the blackout in {Spain}, {Portugal} says about renewables},
  howpublished = {\url{https://www.dw.com/en/spain-portugal-blackout-renewables-wind-solar-energy-grid-v2/a-72606531}},
  year         = {2025},
  note         = {Accessed 08 March 2026}
}

@article{van2025optimizing,
  title={Optimizing power grid topologies with reinforcement learning: A survey of methods and challenges},
  author={van der Sar, Erica and Zocca, Alessandro and Bhulai, Sandjai},
  journal={Foundations and Trends in Electric Energy Systems},
  volume={9},
  number={1},
  pages={1--119},
  year={2025},
  publisher={Emerald Publishing Limited}
}

@article{marot2020learning,
  title={Learning to run a power network challenge for training topology controllers},
  author={Marot, Antoine and Donnot, Benjamin and Romero, Camilo and Donon, Balthazar and Lerousseau, Marvin and Veyrin-Forrer, Luca and Guyon, Isabelle},
  journal={Electric Power Systems Research},
  volume={189},
  pages={106635},
  year={2020},
  publisher={Elsevier}
}

@article{matavalam2022curriculum,
  title={Curriculum based reinforcement learning of grid topology controllers to prevent thermal cascading},
  author={Matavalam, Amarsagar Reddy Ramapuram and Guddanti, Kishan Prudhvi and Weng, Yang and Ajjarapu, Venkataramana},
  journal={IEEE Transactions on Power Systems},
  volume={38},
  number={5},
  pages={4206--4220},
  year={2022},
  publisher={IEEE}
}

@inproceedings{bengio2009curriculum,
  title={Curriculum learning},
  author={Bengio, Yoshua and Louradour, J{\'e}r{\^o}me and Collobert, Ronan and Weston, Jason},
  booktitle={Proceedings of the 26th annual international conference on machine learning},
  pages={41--48},
  year={2009}
}

@inproceedings{lan2020ai,
  title={AI-based autonomous line flow control via topology adjustment for maximizing time-series ATCs},
  author={Lan, Tu and Duan, Jiajun and Zhang, Bei and Shi, Di and Wang, Zhiwei and Diao, Ruisheng and Zhang, Xiaohu},
  booktitle={2020 IEEE Power \& Energy Society General Meeting (PESGM)},
  pages={1--5},
  year={2020},
  organization={IEEE}
}

@article{dorfer2022power,
  author    = {Dorfer, Matthias and Fuxj{\"a}ger, Anton R and Kozak, Kristian and Blies, Patrick M and Wasserer, Marcel},
  title     = {Power grid congestion management via topology optimization with {AlphaZero}},
  journal   = {arXiv preprint arXiv:2211.05612},
  year      = {2022}
}

@article{silver2018general,
  title     = {A general reinforcement learning algorithm that masters chess, shogi, and Go through self-play},
  author    = {Silver, David and Hubert, Thomas and Schrittwieser, Julian and
               Antonoglou, Ioannis and Lai, Matthew and Guez, Arthur and
               Lanctot, Marc and Sifre, Laurent and Kumaran, Dharshan and
               Graepel, Thore and Lillicrap, Timothy and Simonyan, Karen and
               Hassabis, Demis},
  journal   = {Science},
  volume    = {362},
  number    = {6419},
  pages     = {1140--1144},
  year      = {2018},
  publisher = {American Association for the Advancement of Science},
  doi       = {10.1126/science.aar6404}
}

@misc{doornhof2026renewable,
  author       = {Job Doornhof},
  title        = {Renewable curtailment compensation costs in Germany decrease 22\% in 2025},
  howpublished = {\url{https://www.cleanenergywire.org/news/renewable-curtailment-compensation-costs-germany-decrease-22-2025}},
  year         = {2026},
  note         = {Accessed 01 May 2026} 
}

@misc{smard2024,
  author       = {SMARD},
  title        = {The development of congestion management},
  howpublished = {\url{https://www.smard.de/page/en/topic-article/212250/217910/the-development-of-congestion-management}},
  year         = {2024},
  note         = {Accessed 01 May 2026} 
}

@inproceedings{yoon2021winning,
  title={Winning the l2rpn challenge: Power grid management via semi-markov afterstate actor-critic},
  author={Yoon, Deunsol and Hong, Sunghoon and Lee, Byung-Jun and Kim, Kee-Eung},
  booktitle={International Conference on Learning Representations},
  year={2021}
}

@inproceedings{zhou2021action,
  title={Action set based policy optimization for safe power grid management},
  author={Zhou, Bo and Zeng, Hongsheng and Liu, Yuecheng and Li, Kejiao and Wang, Fan and Tian, Hao},
  booktitle={Joint European Conference on Machine Learning and Knowledge Discovery in Databases},
  pages={168--181},
  year={2021},
  organization={Springer}
}

@article{marot2022learning,
  title={Learning to run a power network with trust},
  author={Marot, Antoine and Donnot, Benjamin and Chaouache, Karim and Kelly, Adrian and Huang, Qiuhua and Hossain, Ramij-Raja and Cremer, Jochen L},
  journal={Electric Power Systems Research},
  volume={212},
  pages={108487},
  year={2022},
  publisher={Elsevier}
}

@misc{D3QN2020,
  author       = {Y. Zhihong and others},
  title        = {A winning approach of {NeurIPS} 2020 {L2RPN} Comp},
  howpublished = {GitHub},
  year         = {2020},
  note         = {\url{https://github.com/lujasone/NeurIPS_2020_L2RPN_Comp_An_Approach}, Accessed 10 March 2026}
}

@article{D3QN2022,
  author    = {I. Damjanovic and I. Pavi{\'c} and M. Puljiz and M. Brcic},
  title     = {Deep reinforcement learning-based approach for autonomous power flow control using only topology changes},
  journal   = {Energies},
  volume    = {15},
  number    = {19},
  pages     = {6920},
  year      = {2022}
}

@misc{lajavanessrepo,
  author       = {{La Javaness}},
  title        = {{L2RPN} with {PPO}: A Winning Solution for the {L2RPN} Challenge},
  howpublished = {GitHub},
  year         = {2026},
  note         = {\url{https://github.com/lajavaness/l2rpn-with-ppo}, Accessed 10 March 2026}
}

@misc{l2rpn_baselines,
  author       = {{RTE-France}},
  title        = {{L2RPN} Baselines: A repository to host baselines for {L2RPN} competitions},
  howpublished = {GitHub},
  year         = {2026},
  note         = {\url{https://github.com/rte-france/l2rpn-baselines}, Accessed 22 March 2026}
}

@misc{sintes2024how,
  author       = {Sintes, J. and Dang, V. T.},
  title        = {{How we built the winning real-time autonomous agent for power grid management in the {L2RPN} challenge 2023}},
  howpublished = {Medium},
  year         = {2024},
  note         = {\url{https://lajavaness.medium.com/how-we-built-the-winning-real-time-autonomous-agent-for-power-grid-management-in-the-l2rpn-41ab3cfaddbd}, Accessed 08 May 2026}
}

@misc{martinez2021l2rpn,
  author       = {Martinez, H.},
  title        = {{L2RPN}},
  howpublished = {GitHub},
  year         = {2021},
  note         = {\url{https://github.com/horacioMartinez/L2RPN}, Accessed 08 May 2026}
}

@misc{lair2023artificial,
  author       = {Lair, N. and Bossavy, A. and Champion, P. and Renault, V.},
  title        = {{Artificial agents designed to run a power network - White Paper}},
  howpublished = {Artelys},
  year         = {2023},
  note         = {\url{https://www.artelys.com/app/uploads/2024/04/White_paper_L2RPN_2023_.pdf}, Accessed 08 May 2026}
}

@article{lehna2023managing,
  author    = {Lehna, Malte and Viebahn, Jan and Marot, Antoine and Tomforde, Sven and Scholz, Christoph},
  title     = {Managing power grids through topology actions: A comparative study between advanced rule-based and reinforcement learning agents},
  journal   = {Energy and AI},
  volume    = {14},
  pages     = {100276},
  year      = {2023},
  publisher={Elsevier}
}

@misc{delft2023l2rpn,
  author       = {{D. A. E. Lab}},
  title        = {{Learning to Run a Power Network - Delft 2023 Competition}},
  howpublished = {\url{https://codalab.lisn.upsaclay.fr/competitions/12420}},
  note         = {CodaLab; Accessed 08 May 2026},
  year         = {2023}
}

@misc{huawei2020l2rpn,
  author       = {{H. T. EI Innovation Lab, Huawei Cloud}},
  title        = {{NeurIPS Competition 2020: Learning to Run a Power Network (L2RPN) - Robustness Track}},
  howpublished = {\url{https://github.com/leizhu0608/L2RPN-2020-Robustness-Track-Solution}},
  note         = {Accessed 08 May 2025},
  year         = {2020}
}

@inproceedings{chauhan2023powrl,
  author    = {Chauhan, Anandsingh and Baranwal, Mayank and Basumatary, Ansuma},
  title     = {{PowRL}: A reinforcement learning framework for robust management of power networks},
  booktitle = {Proceedings of the AAAI Conference on Artificial Intelligence},
  volume    = {37},
  number={12},
  pages     = {14757--14764},
  year      = {2023}
}

@article{manczak2023hierarchical,
  author    = {Manczak, Blazej and Viebahn, Jan and van Hoof, Herke},
  title     = {Hierarchical reinforcement learning for power network topology control},
  journal   = {arXiv preprint arXiv:2311.02129},
  year      = {2023}
}

@inproceedings{marot2021learning,
  title={Learning to run a power network challenge: a retrospective analysis},
  author={Marot, Antoine and Donnot, Benjamin and Dulac-Arnold, Gabriel and Kelly, Adrian and O’Sullivan, Aidan and Viebahn, Jan and Awad, Mariette and Guyon, Isabelle and Panciatici, Patrick and Romero, Camilo},
  booktitle={NeurIPS 2020 competition and demonstration track},
  pages={112--132},
  year={2021},
  organization={PMLR}
}

@inproceedings{butt2024reinforcement,
  title={Why reinforcement learning in energy systems needs explanations},
  author={Butt, Hallah Shahid and Schafer, Benjamin},
  booktitle={Proceedings of the 2024 Workshop on Explainability Engineering},
  pages={26--30},
  year={2024}
}

@inproceedings{butt2024explainable,
  title={Explainable Reinforcement Learning for Optimizing Electricity Costs in Building Energy Management},
  author={Butt, Hallah Shahid and Huang, Qiong and Sch{\"a}fer, Benjamin},
  booktitle={2024 3rd International Conference on Energy Transition in the Mediterranean Area (SyNERGY MED)},
  pages={1--6},
  year={2024},
  organization={IEEE}
}

@inproceedings{subramanian2021exploring,
  title={Exploring grid topology reconfiguration using a simple deep reinforcement learning approach},
  author={Subramanian, Medha and Viebahn, Jan and Tindemans, Simon H and Donnot, Benjamin and Marot, Antoine},
  booktitle={2021 IEEE Madrid PowerTech},
  pages={1--6},
  year={2021},
  organization={IEEE}
}

@article{wolgast2024learning,
  title={Learning the optimal power flow: Environment design matters},
  author={Wolgast, Thomas and Nie{\ss}e, Astrid},
  journal={Energy and AI},
  volume={17},
  pages={100410},
  year={2024},
  publisher={Elsevier}
}

% \appendix
\appendix
\section{Appendix}
\label{app:}
\subsection{Curriculum Learning (TTJS) Framework} \label{app:ttjs}
Using the same setup and hyperparameters as PPO Topology, we supplemented it with an additional Curriculum Learning setup as provided by the La Javaness \cite{sintes2024how} combined with the Curriculum Agent \cite{lehna2023managing} repository. These apply a Teacher-Tutor-Junior (TTJS) framework where the ``Teachers'' run a brute-force search to collect a large amount of trajectories (approximately 50GB). The ``Tutors'' then filter these for high-quality experiences to reduce the action space accordingly and initialize the Junior neural network via IL. Only after this, does the Senior apply the actual RL utilizing the previously initialized neural network.

\subsection{MCTS Mathematical Framework and Structure} \label{app:mcts_math}
Whenever the AlphaZero agent reaches a state with a maximin line load that is above our activation threshold, the iterative MCTS algorithm is triggered. In each iteration, we traverse the tree from the root, which represents the current state of the power grid, until we reach the leaf layer, where a new node is then appended. The traversal follows the Predictor Upper Confidence Bound applied to Trees (PUCT) formula:
\begin{equation}
a_t = \text{argmax}_a \left( Q(s,a) + c_{puct} \cdot P(s,a) \cdot \frac{\sqrt{\sum_b N(s,b)}}{1 + N(s,a)} \right),
\end{equation}
where:
\begin{itemize}
  \item $Q(s,a)$ is the action-value (the mean reward of choosing action $a$ in state $s$).
  \item $P(s,a)$ is the prior probability of choosing action $a$ as provided by the policy network.
  \item $N(s,a)$ is the visit count for the specific edge, while \\$\sum_b N(s,b)$ is the total visit count of the parent node.
  \item $c_{puct}$ is a constant that controls the level of exploration.
\end{itemize}

\cite{dorfer2022power} suggests determining the value of the leaf nodes added to the MCTS using the heuristic function:
\begin{equation} \label{eq2}
    v_t = \sum_{j=t}^{t+h} \gamma^{j-t} r_j
\end{equation}
where $h$ is a sufficiently large horizon depending on the discount factor $\gamma$ \cite{silver2018general}, and the reward $r_j$ at the current time step $j$. To manage computational complexity, an early stopping mechanism is included. This mechanism terminates the search once a specific number of recovery nodes are identified. Specifically, when $t_{skipped}$ steps are skipped between critical nodes, a ``recovery node'' is added to mark an action that solved the congestion and led to a period of grid stability. The search is interrupted once $t_{\text{stopping}}$ recovery nodes have been identified or we have exhausted a predetermined budget of simulations. The training target is then determined by evaluating which branch of the tree reached the greatest depth in terms of survival steps. The action corresponding to this deepest subtree is selected as the training target for the current root state. These state-action pairs are appended to a replay buffer, which is used to update the neural network once a sufficient number of chronics have been processed to ensure a diverse and representative dataset.

% reward
\subsection{MCTS Reward Function}
\label{app:reward} 
Previous studies have employed a wide variety of reward functions. In this work, we cover the most relevant categories, including safety margins, maximum line loading, survival time, and energy efficiency, as well as composite multi-objective strategies. The full mathematical definitions for the AlphaZero Reward, D3QN-2022 Survival Reward, D3QN-2020 Composite Reward, Loss Reward, MaxRho Reward, and PPO Reward are summarized in \cite{van2025optimizing} and given as follows:

\begin{itemize}
    \item \textbf{AlphaZero Reward} \cite{dorfer2022power}
    Originally utilized with the AlphaZero agent, this reward is designed to penalize high line loading and network fragmentation. It uses an exponential decay function to provide a smooth gradient for the MCTS:
    \begin{equation}
        R_{\text{AlphaZero}} = e^{-u(t) - 0.5 \cdot n_{\text{offline}}}
    \end{equation}
   The penalty term $u(t)$ is calculated based on the maximum line loading ratio $\rho_{\max}$:
\begin{equation}
u(t) = 
\begin{cases} 
    \max(\rho_{\max} - 0.5, 0) & \text{if } \rho_{\max} \leq 1 \\
    \sum_{i \in \mathcal{O}} (\rho_i - 0.5) & \text{if } \rho_{\max} > 1
\end{cases}
\end{equation}
where $\mathcal{O}$ denotes the set of overflowing lines (i.e., all lines $i$ for which $\rho_i > 1$). The term $n_{\text{offline}}$ accounts for the number of lines currently disconnected.

    \item \textbf{D3QN-2022 Survival Reward} \cite{D3QN2022}
    A binary reward signal focusing exclusively on survival:
    \begin{equation}
        R_{\text{D3QN-2022}} =
        \begin{cases}
            1 & \text{if grid is operational at step } t \\
            0 & \text{if game over or illegal action} 
        \end{cases}
    \end{equation} 

    \item \textbf{D3QN-2020 Composite Reward} \cite{D3QN2020}\\
    This is a multi-objective reward function that balances grid stability, safety margins, and topological similarity:
    \begin{equation}
\begin{split}
\begin{aligned}
    R_{\text{D3QN-2020}} = & \ w_1 \cdot R_{\text{sandbox}} + w_2 \cdot R_{\text{overflow}} \\
                           & + w_3 \cdot R_{\text{distance}} + w_4 \cdot R_{\text{capacity}}
\end{aligned}
\end{split}
\end{equation}
    The weights are $w_1$=30, $w_2$=200, $w_3$=20, $w_4$=3. $R_{\text{sandbox}}(t)$ is based on the costs for redispatch and power loss, while $R_{\text{overflow}}(t)$ rewards a lower magnitude of line overflows. $R_{\text{distance}}(t)$ is determined by the distance from the reference topology (where a smaller distance yields a higher reward), and $R_{\text{capacity}}(t)$ rewards global safety margins based on the available line capacity.

    \item \textbf{Loss Reward (Efficiency-Based)} \cite{yoon2021winning}
    Targets the minimization of active power losses:
    \begin{equation}
        R_{\text{Loss}}(t) = \frac{\sum P_{\text{loads}}}{\sum P_{\text{gens}}} - 0.9
    \end{equation} 

    \item \textbf{MaxRho Reward (LJN Agent)} \cite{lajavanessrepo}
    % \cite{lajavanessrepo} cannot find in internet
    A linear reward aimed at maximizing the safety margin:
    \begin{equation}
        R_{\text{MaxRho}}(t) = 2.0 - \rho_{\text{max}}
    \end{equation} 

    \item \textbf{PPO Reward (LJN Agent)} \cite{lajavanessrepo}
    Combines the MaxRho principle with a progressive step-penalty:
    \begin{equation}
        R_{\text{PPO}}(t) = (2.0 - \rho_{\text{max}}) - \text{penalty} + \text{bonus}_{\text{action}}
    \end{equation}
    The \textit{penalty} term is defined as:
    \begin{equation}
        \text{penalty} =
        \begin{cases}
            0 & \text{if } \rho_{\text{max}} \le 0.8 \\
            1 & \text{if } 0.8 < \rho_{\text{max}} \le 0.9 \\
            2 & \text{if } 0.9 < \rho_{\text{max}} \le 0.95 \\
            3 & \text{if } 0.95 < \rho_{\text{max}} \le 1.0 \\
            5 & \text{if } \rho_{\text{max}} > 1.0
        \end{cases}
    \end{equation}
    A $\text{bonus}_{\text{action}} = 0.1$ is applied for the ``Do Nothing'' action.
\end{itemize}
% Appendix \ref{app:rewards_full}.

\subsection{MCTS Guidance Variants: Detailed Descriptions} \label{app:guidance_variants}
To optimize the search efficiency of the AlphaZero agent, we examine three distinct methods for guiding the MCTS process: 

\begin{itemize}
    \item \textbf{IL Variant:} This configuration operates without using the learned policy or a value function within the MCTS. It relies exclusively on the search depth and the transition rewards encountered during tree expansion to identify surviving trajectories, essentially treating the MCTS as systematic discovery mechanism.
    \item \textbf{Original Heuristic Approach:} This version uses the policy prior to narrow the search breadth by focusing on high-probability actions. Additionally, it employs the heuristic value function from \autoref{eq2} to estimate the quality of newly added leaf nodes.
    \item \textbf{Learned Q-Function:} In this setup, node selection and evaluation are driven by a combination of the prior policy and learned Q-values. By replacing the heuristic estimation with predictions by the neural network, the agent attempts to leverage learned experience to prioritize more promising search branches.
\end{itemize}

\section{MCTS}
\subsection{MCTS Hyperparameter Configurations} \label{app:hyp_configs}
To investigate the impact of search depth and efficiency, the following configurations for Early Stopping and Horizon size were evaluated:
\paragraph{Early Stopping Variants}
\begin{itemize}
    \item \textbf{Variant 1:} $t_{\text{skipped}} = 200, t_{\text{stopping}} = 50$
    \item \textbf{Variant 2:} $t_{\text{skipped}} = 200, t_{\text{stopping}} = 20$
    \item \textbf{Variant 3:} $t_{\text{skipped}} = 50, t_{\text{stopping}} = 10$
    \item \textbf{Variant 4:} $t_{\text{skipped}} = 10, t_{\text{stopping}} = 5$
\end{itemize}

\paragraph{Horizon Size and Reset Thresholds}
We compared lookahead horizons of 20, 100, 200, and 500 steps applied to \autoref{eq2}. Additionally, the topology reset threshold was examined across values of 75\%, $80\%$, $85\%$, $90\%$, and $95\%$ defined by the maximum line load observed within the grid.
\begin{figure}[ht]
  \centering
  \includegraphics[width=\linewidth]{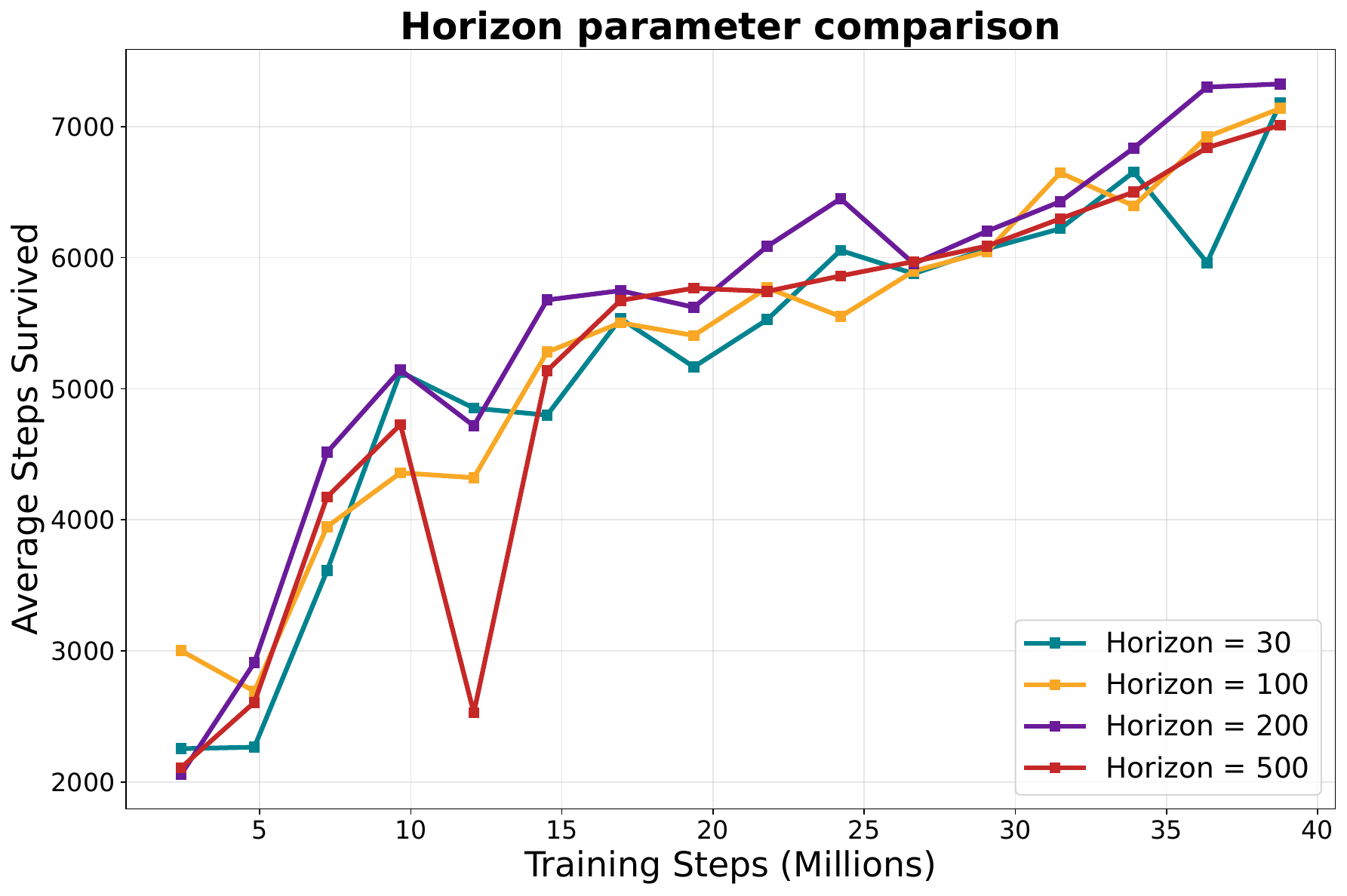}
  \caption{Evaluation of different horizon sizes for the heuristic based AlphaZero baseline approach.} 
  \label{obser_horizon}
  \Description{}
\end{figure}

\begin{figure}[ht]
  \centering
  \includegraphics[width=\linewidth]{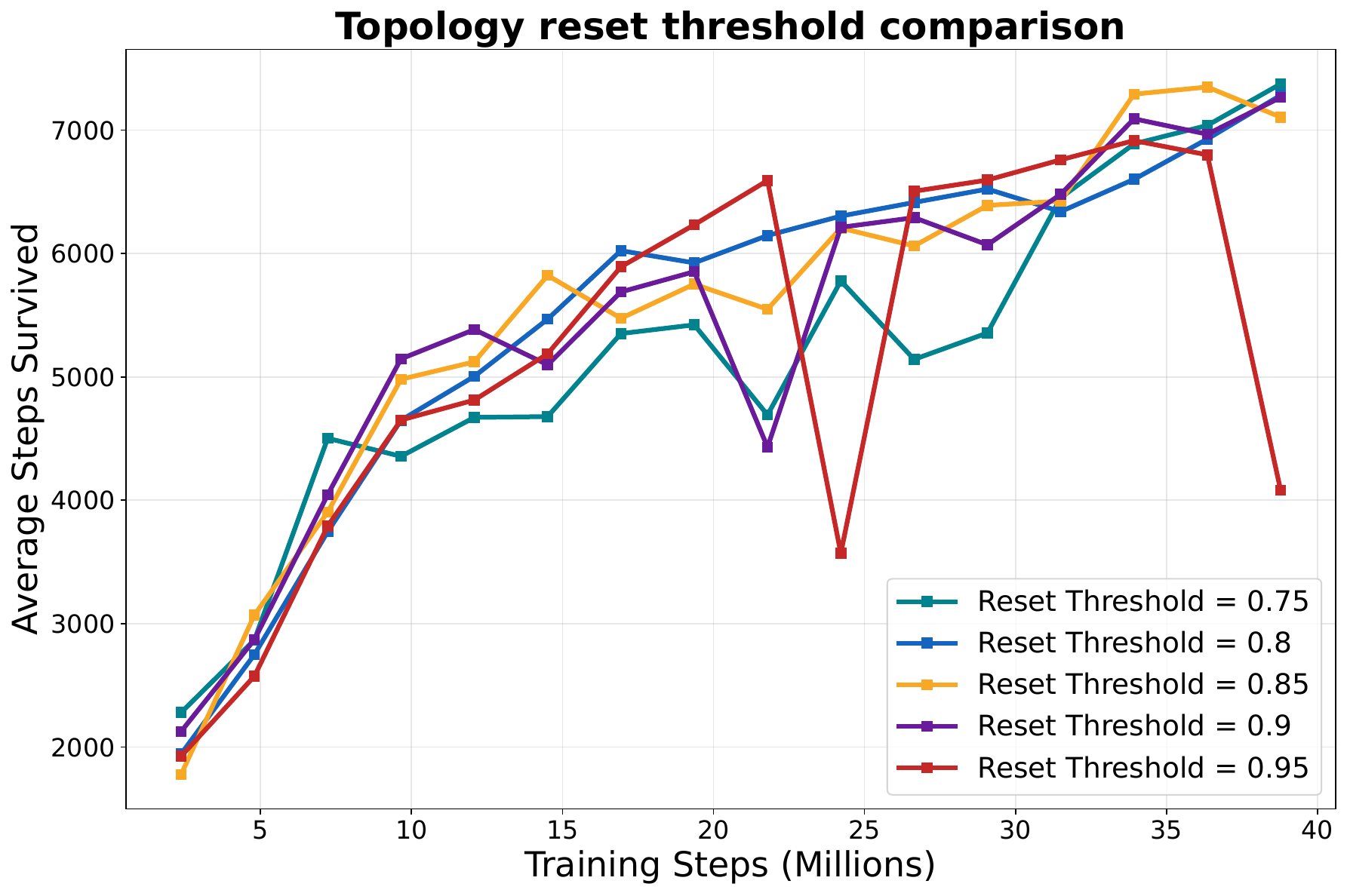}
  \caption{Evaluation of different thresholds for the application of the heuristic that leads the agent back to the reference topology.} 
  \label{obser_threshold}
  \Description{}
\end{figure}

\begin{figure}[ht] %\label{safetyparams}
  \centering
  \includegraphics[width=\linewidth]{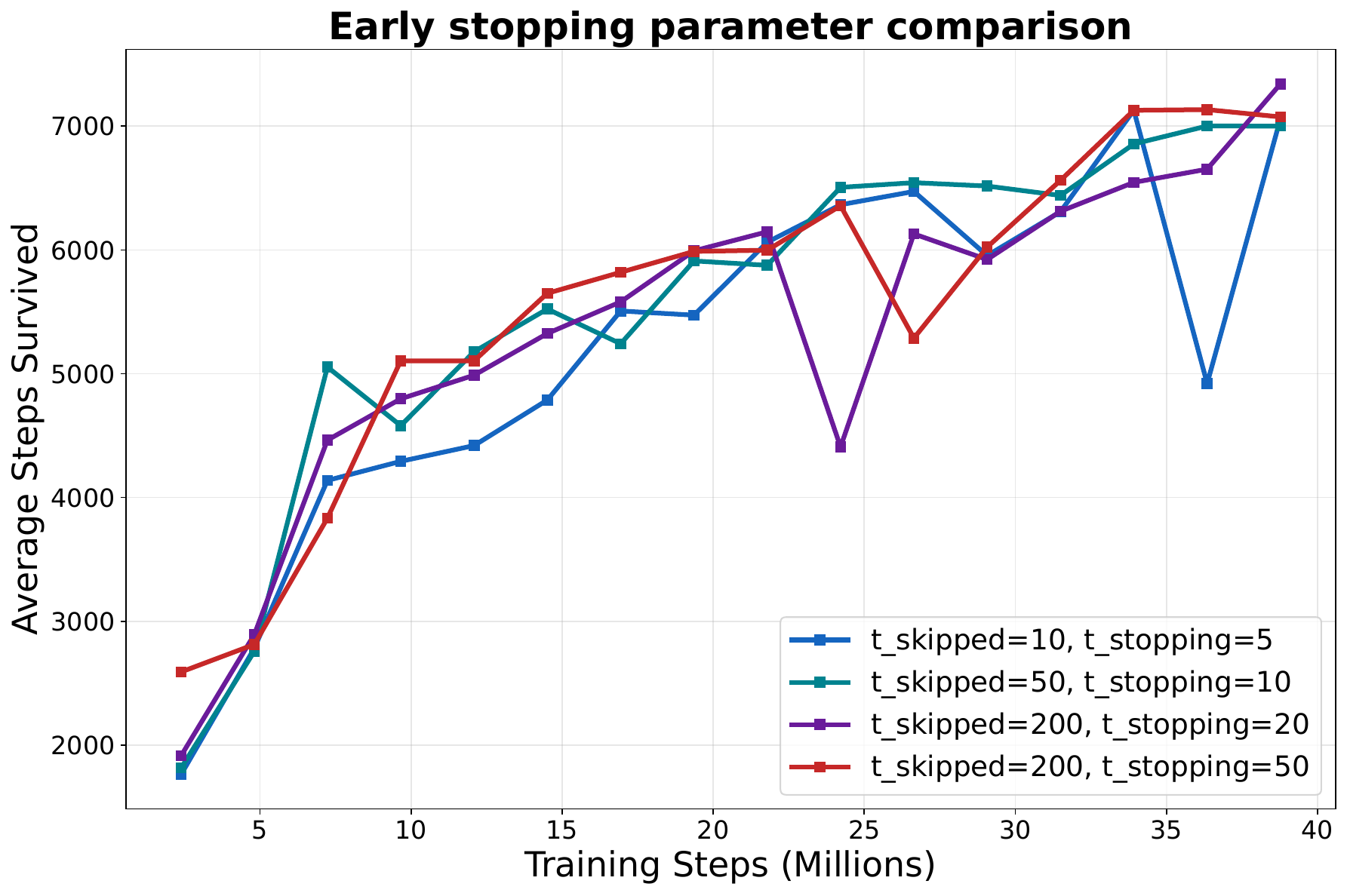}
  \caption{Evaluation of values for the $t_{stopping}$ and $t_{skipped}$ hyperparameters which define the early stopping mechanism for the MCTS tree.} 
  \label{obser_hyper}
  \Description{}
\end{figure}

\begin{figure}[ht]
  \centering
  \includegraphics[width=\linewidth]{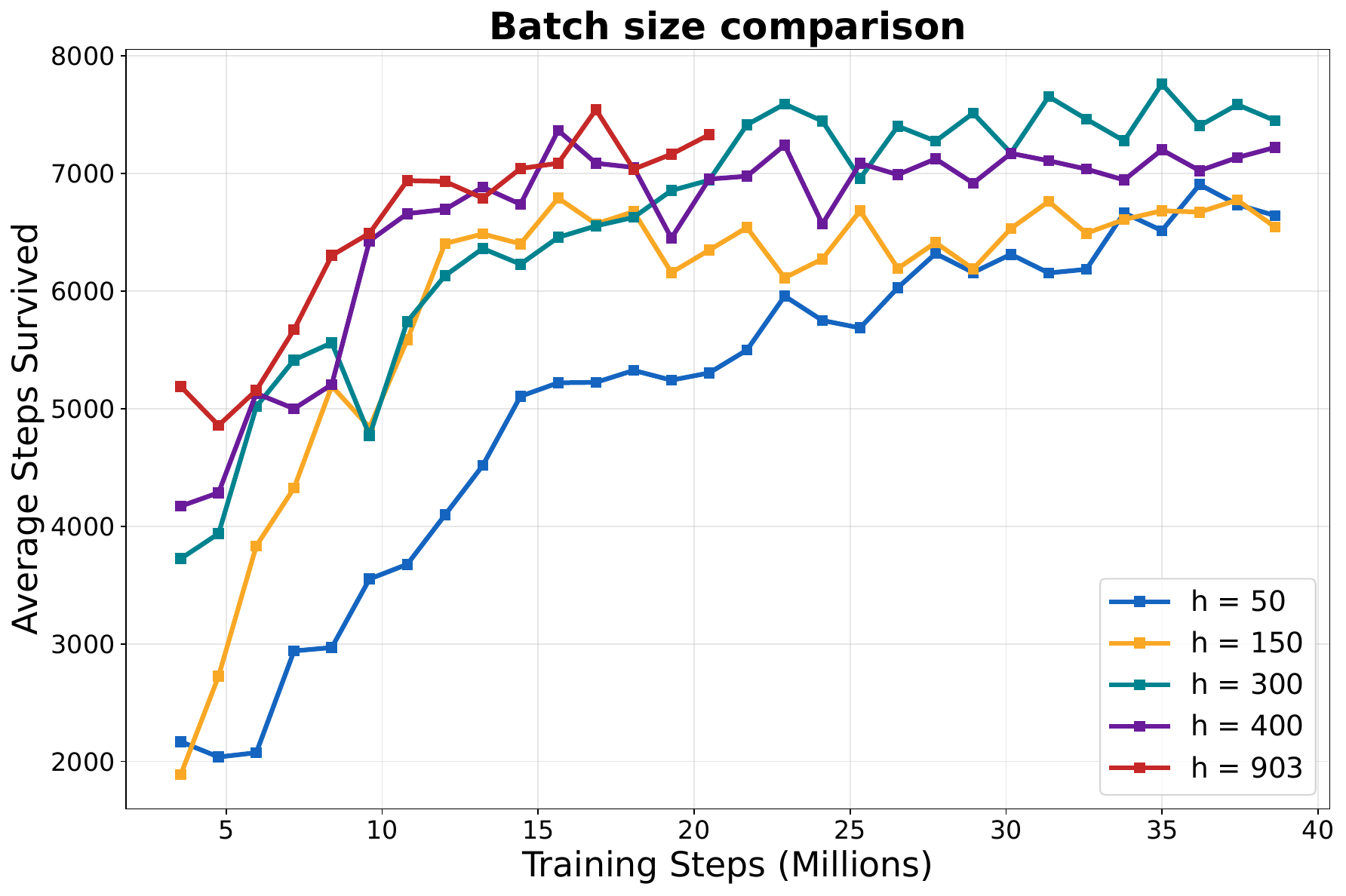}
  \caption{Evaluation of the effect caused by different choices for the batch size which determines the number of chronics we collect data from between training iterations.} 
  \label{obser_batchsize}
  \Description{}
\end{figure}

\subsubsection{MCTS Hyperparameters}
The sensitivity of MCTS hyperparameters reveals a consistent trend: while final agent performance is relatively robust across configurations, parameter selection significantly impacts the stability and efficiency of the training process. Overall, these results indicate that the primary role of MCTS hyperparameters tuning is not to lift the performance ceiling, but rather to enhance computational efficiency and training robustness. 

\paragraph{Search and Stopping Criteria}
As shown in \autoref{obser_hyper}%{safetyparams}
, the medium early stopping configuration ($t_{skipped}$ = 50 and $t_{stopping}$ = 10) avoided the training outliers seen in more aggressive or conservative variants. An exception can be seen in the batch size experiment, where we learn that smaller batch sizes can lead to worse performance. Our evaluation of early-stopping parameters suggests that aggressive stopping criteria can be adopted to significantly reduce computational overhead without degrading final survival times. 
% First, the results for the early-stopping parameters $t_{\text{skipped}}$ and $t_{\text{stopping}}$ suggest that lower values can be adopted to significantly reduce computational overhead.
Since the final survival times did not degrade under more aggressive early-stopping criteria, these settings offer a clear path to increasing the real-time feasibility of MCTS-based controllers. However, the superior stability of the ``medium'' configuration (Variant 2) indicates that some degree of search depth is necessary to prevent the high-variance training spikes observed in the more minimal variants. %\newline

\paragraph{Horizon and Thresholds}
Regarding look-ahead depth, shorter horizons facilitated a more consistent learning trajectory. The longest horizon (500 steps) exhibited the pronounced instability, likely due to the diminishing validity of stationary reward assumptions over extended durations as grid dynamics evolve and stochasticity increases. Furthermore, a topology reset threshold of 0.85 emerged as the most robust setting. Higher thresholds, which delay intervention until the grid is more severely stressed, resulted in more frequent performance ``down-spikes'', suggesting that overly delayed interventions lead to more volatile grid states.

% This behavior likely stems from the diminishing validity of the stationary reward assumption over extended durations, which fails to encapsulate grid dynamics. However, as noted by \cite{dorfer2022power}, the optimal horizon length may depend on the chosen discount factor.
\paragraph{Batch Size Efficiency}
Finally, the batch size experiments demonstrate that insufficiently small batches lead to degraded performance, likely due to a lack of gradient diversity. While performance remained comparable across larger scales, the medium-sized batches provided the optimal balance between wall-clock training time and convergence stability.

% the highest stability observed at a 0.85 topology reset threshold further suggest that overly aggressive interventions can be counter-productive and lead to a more volatile and unstable grid. Overall, these results ultimately indicate that the primary role of MCTS hyperparameter tuning is not to lift the performance ceiling, but rather to enhance computational efficiency and training robustness.

\subsection{Detailed Baseline Configuration} \label{app:baseline_details}
To compare all of the previously mentioned AlphaZero design options we will use a baseline configuration with the following settings:
\begin{itemize}
  \item \textbf{Activation Threshold:} 0.98 (max. line load of $98\%$).
  \item \textbf{Search Limits:} Maximum of 250 MCTS simulations per step.
  \item \textbf{Temporal Parameters:} $t_{skipped}$ = 200 and $t_{stopping}$ = 50.
  \item \textbf{Reward Function:} AlphaZero reward.
  \item \textbf{Safety Heuristics:} Automatic reconnection of a disconnected line after coldown and reversion to the initial reference topology when in safe state (below 75\% max. line load).
  \item \textbf{Observation Space:} Minimal - restricted to line loads only.
\end{itemize}

\end{document}